\pdfoutput=1

\documentclass[11pt]{article}

\usepackage[final]{acl}

\usepackage{times}
\usepackage{latexsym}

\usepackage[T1]{fontenc}

\usepackage[utf8]{inputenc}

\usepackage{microtype}

\usepackage{inconsolata}

\usepackage{graphicx}

\usepackage{soul}
\usepackage{url}
\usepackage{amsthm}
\usepackage{booktabs}
\usepackage{algorithm}
\usepackage{algorithmic}
\usepackage[switch]{lineno}

\usepackage{enumitem}
\setlist[itemize]{nosep,leftmargin=*}

\usepackage[T1]{fontenc}    
\usepackage{nicefrac}       
\usepackage{microtype}      
\usepackage{xcolor}         
\usepackage{multicol}
\usepackage{bbding}
\usepackage{color,colortbl}
\usepackage[export]{adjustbox}
\usepackage{array}

\definecolor{myorange}{RGB}{251, 229, 214}
\definecolor{lightred}{RGB}{251,49,153}

\usepackage{pifont} 
\usepackage{hyperref}
\hypersetup{colorlinks,linkcolor={red}} 
\usepackage{bibentry}
\usepackage{multirow}
\usepackage{amsmath,amssymb,amsfonts}
\usepackage[most]{tcolorbox}
\newtcolorbox{promptbox}{
colback=gray!5,  
colframe=black!75, 
left=1em, 
right=1em, 
top=1em, 
bottom=1em, 
sharp corners, 
boxrule=1pt 
}

\usepackage{booktabs}
\usepackage{makecell}
\usepackage{graphicx}
\usepackage{tikz}
\usepackage{array}

\usepackage{float}
\usepackage{arydshln}
\usepackage{enumitem}
\usepackage{hyperref}
\usepackage{multirow}

\newcommand\diag[4]{%
  \multicolumn{1}{p{0.8cm}}{\hskip-\tabcolsep
  $\vcenter{\begin{tikzpicture}[baseline=0,anchor=south west,inner sep=#1, scale=0.85]
  \path[use as bounding box] (0,0) rectangle (#2+0.5\tabcolsep,1.8\baselineskip); 
  \node[minimum width={#2+0.5\tabcolsep-\pgflinewidth},
        minimum height=1.5\baselineskip] (box) {};
  \draw[line cap=round] (box.north west) -- (box.south east);
  \node[anchor=south west, font=\tiny] at ([yshift=-0.5ex]box.south west) {#3}; 
  \node[anchor=north east, font=\tiny] at ([yshift=0.2ex]box.north east) {#4}; 
 \end{tikzpicture}}$\hskip-\tabcolsep}}

\usepackage{caption}

\title{Can LLMs Improve Multimodal Fact-Checking by Asking Relevant Questions?
}

\author{
 Alimohammad Beigi\textsuperscript{1}\quad 
 Bohan Jiang\textsuperscript{1}\quad 
 Dawei Li\textsuperscript{1}\quad 
 Zhen Tan\textsuperscript{1}\quad 
\\
 \textbf{Pouya Shaeri\textsuperscript{1}}\quad 
 \textbf{Tharindu Kumarage\textsuperscript{1}}\quad 
 \textbf{Amrita Bhattacharjee\textsuperscript{1}}\quad 
 \textbf{Huan Liu \textsuperscript{1}}\quad 
\\
\\
 \textsuperscript{1}School of Computing, and Augmented Intelligence, Arizona State University
\\
 {\tt \{abeigi, bjiang14, daweili5, ztan36, pshaeri, kskumara, abhatt43, huanliu\}@asu.edu}\\
}

\begin{document}
\maketitle
\begin{abstract}
Traditional fact-checking relies on humans to formulate relevant and targeted fact-checking \textit{questions} (FCQs), search for evidence, and verify the factuality of claims. While Large Language Models (LLMs) have been commonly used to automate evidence retrieval and factuality verification at scale, their effectiveness for fact-checking is hindered by the absence of FCQ formulation. 
To bridge this gap, we seek to answer two research questions: (1) Can LLMs generate relevant FCQs? (2) Can LLM-generated FCQs improve multimodal fact-checking?
We therefore introduce a framework \textsc{Lrq-Fact} for using LLMs to generate relevant FCQs  to facilitate evidence retrieval and enhance fact-checking by probing information across multiple modalities.
Through extensive experiments, we verify if \textsc{Lrq-Fact} can generate relevant FCQs of different types and if \textsc{Lrq-Fact} can consistently outperform baseline methods in multimodal fact-checking. Further analysis illustrates how each component in \textsc{Lrq-Fact} works toward improving the fact-checking performance.
\end{abstract}

\section{Introduction}

Fact-checking is an important yet challenging task in combating online misinformation. Modern misinformation often spreads across multiple modalities, containing both textual and visual falsehoods, which significantly complicates accurate and efficient fact-checking~\cite{akhtar2023multimodal}. In journalism, fact-checking is traditionally a three-step process~\cite{graves2019fact}, where human fact-checkers (1) formulate relevant and targeted fact-checking \textit{questions} (FCQs), (2) search for supporting evidence, and (3) verify the factuality of claims or statements. Fact-checkers leverage domain knowledge to pose precise and contextually relevant FCQs, ensuring claims are evaluated from multiple perspectives~\cite{politifact2011, vlachos2014fact, vo2019learning}. However, given the rapid proliferation of online misinformation~\cite{chen2023can, jiang2024catching}, manual fact-checking is insufficient to keep pace with the scale of the problem~\cite{shaeri2023semi}.

\begin{figure}[t]
	\centering
	\includegraphics[width=1\columnwidth]{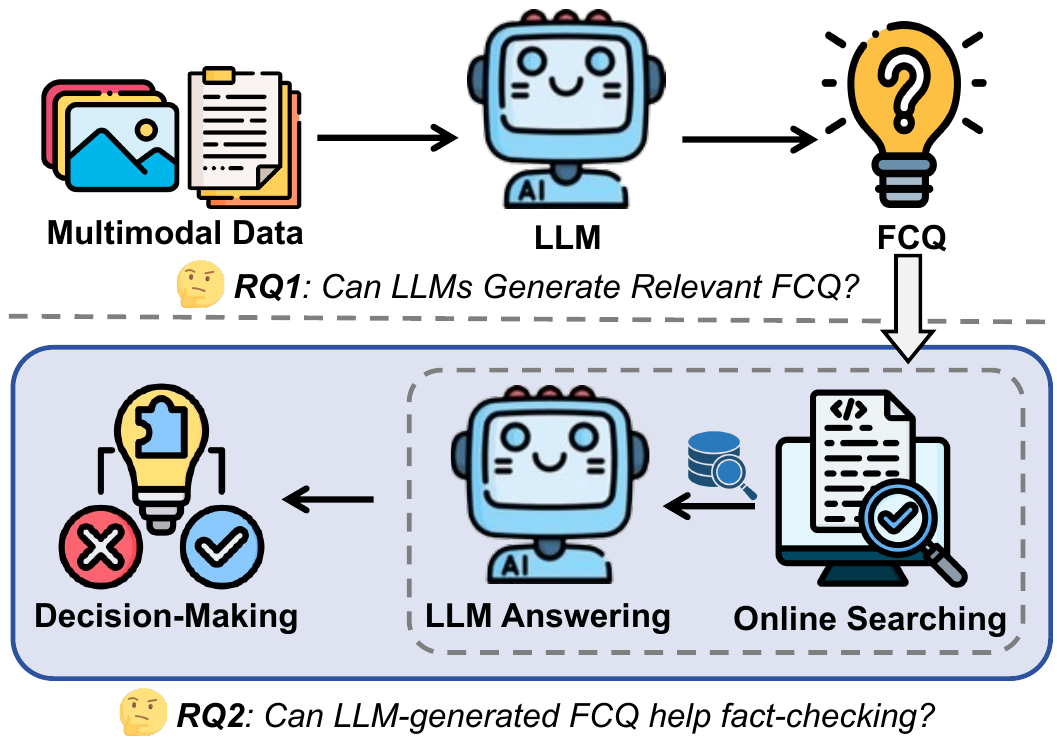}
	\caption{The two research questions we aim to address in this work.}
	\label{fig:RQs}
	\vspace{-8mm}
\end{figure}

To improve efficiency, researchers have developed automated fact-checking (AFC) systems capable of identifying misinformation~\cite{hassan2017toward, miranda2019automated, dierickx2023automated}. Recently, Large Language Models (LLMs) have been explored for zero-shot fact-checking~\cite{geng2024multimodal}. However, research pointed out that directly prompting LLMs for fact-checking remains less effective in many cases~\cite{yao2023end}. One key issue is the absence of relevant FCQs, which are essential to guide LLMs in retrieving accurate supporting evidence and conducting reliable veracity evaluations~\cite{chen2022generating, pan2023qacheck, setty2024surprising}.
In this work, we investigate the potential of LLM-generated FCQs
by answering two research questions (Figure~\ref{fig:RQs}): 
\begin{itemize}
    \item Are LLMs capable of generating relevant FCQs?
    \item Can the generated FCQs improve AFC systems? 
\end{itemize}

Inspired by the human fact-checking process, we introduce \underline{\textbf{L}}LM-generated \underline{\textbf{R}}elevant \underline{\textbf{Q}}uestions for multimodal \underline{\textbf{FACT}}-checking (\textsc{Lrq-Fact}), an LLM-based framework designed to automatically generate relevant and targeted FCQs to guide the AFC system to fact-check multimodal misinformation. \textsc{Lrq-Fact} first generates two types of FCQs: (1) visual FCQs, which assess whether an image accurately represents critical details such as people, objects, or events mentioned in the text, and (2) textual FCQs, which question whether the textual claims or statements are supported by evidence. Human annotators and LLM judges evaluate the quality of LLM-generated FCQs with pre-defined rules, and show that most of the textual and visual FCQs generated by \textsc{Lrq-Fact} are contextually relevant to the fact-checking task.

Next, we seek to answer whether the generated questions can improve multimodal fact-checking. With the guidance of relevant FCQs, \textsc{Lrq-Fact} incorporates the internal training knowledge of LLM and external online searching to strengthen its evidence retrieval and verification capabilities. The up-to-date online information is particularly valuable when fact-checking claims related to emerging or rapidly evolving events, where LLMs often lack sufficient ground truth knowledge.
Extensive experiments are conducted on three datasets. Our results show that \textsc{Lrq-Fact} can outperform baseline methods significantly. Furthermore, our ablation study probes into the model's modular components to evaluate their effectiveness in fact-checking performance. Last but not least, we demonstrate that \textsc{Lrq-Fact} is highly adaptive, generalizing across different LLM backbones. In summary, the key contributions of this work are as follows:
\begin{itemize}
    \item We analyze the use of LLMs to generate relevant and targeted FCQs.
    \item We explore and experiment the effectiveness of LLM-generated FCQs in multimodal fact-checking on three benchmark datasets.
    \item Further analysis illustrates how each component in \textsc{Lrq-Fact} contributes to performance.
\end{itemize}

\section{Related Work}

\subsection{Multimodal Misinformation}
Misinformation spans multiple domains, consisting of various modalities such as text and images, making detection increasingly complex~\cite{li2020mm, jiang2024disinformation, tufchi2023comprehensive}.
While early misinformation detection mainly focused on textural content~\cite{thorne2018fever, shu2020fakenewsnet}, recent datasets have incorporated multimodal misinformation, such as Fakeddit~\cite{nakamura2019r}, DGM$^4$~\cite{shao2023detecting}, and MMFakeBench~\cite{liu2024mmfakebench}. Multimodal misinformation detection methods have been developed to learn joint representations of different modalities~\cite{abdali2022multi, zhou2020unified}. Some studies proposed explainable detection frameworks to enhance the interpretability~\cite{liu2023interpretable, fung2021infosurgeon}. However, these models may fall short when facing newly emerged or rapidly evolving topics. Our method addresses this issue by utilizing up-to-date Google Search as an external knowledge source.

\subsection{Fact-Checking}
Fact-checking is essential for combating misinformation, traditionally relying on human fact-checkers to verify claims by generating FCQs and cross-referencing credible sources~\cite{graves2018understanding, graves2019fact}. However, manual fact-checking is resource-intensive and struggles to scale with increasing online misinformation~\cite{vlachos2014fact}. AFC systems address this challenge, with early approaches focusing on textual claims using Machine Learning and Natural Language Processing techniques~\cite{guo2022survey, nakov2021automated, karadzhov2017fully}. Recent studies have expanded AFC's capabilities by integrating large-scale evidence retrieval~\cite{nie2019combining, akhtar2023multimodal, geng2024multimodal}. However, the lack of FCQs limited the performance of AFC. We address this limitation by harnessing LLMs to generate relevant and targeted FCQs.

\subsection{Language Models for Fact-Checking}
LLMs and Vision Language Models (VLMs) have had significant impacts on AFC. LLMs from the GPT family~\cite{brown2020language, achiam2023gpt, hurst2024gpt} and LLaMA series~\cite{touvron2023llama, vavekanand2024llama} excel in language understanding and context-aware question generation, enhancing AFC performance~\cite{achiam2023gpt, vavekanand2024llama, beigi2024model}. VLMs such as CLIP~\cite{radford2021learning}, ViLBERT~\cite{lu2019vilbert}, and Paligemma~\cite{beyer2024paligemma}, enable cross-modal analysis, aligning visual and textual features to detect false statements. Recent studies highlight the potential of combining LLMs and VLMs to generate targeted FCQs for better fact-checking~\cite{singh2021detecting, chen2022generating, pan2023qacheck}. Our framework utilizes LLMs and VLMs to generate relevant FCQs, improving multi-modal fact-checking.

\begin{figure*}[h]
	\centering
	\includegraphics[width=1\textwidth]{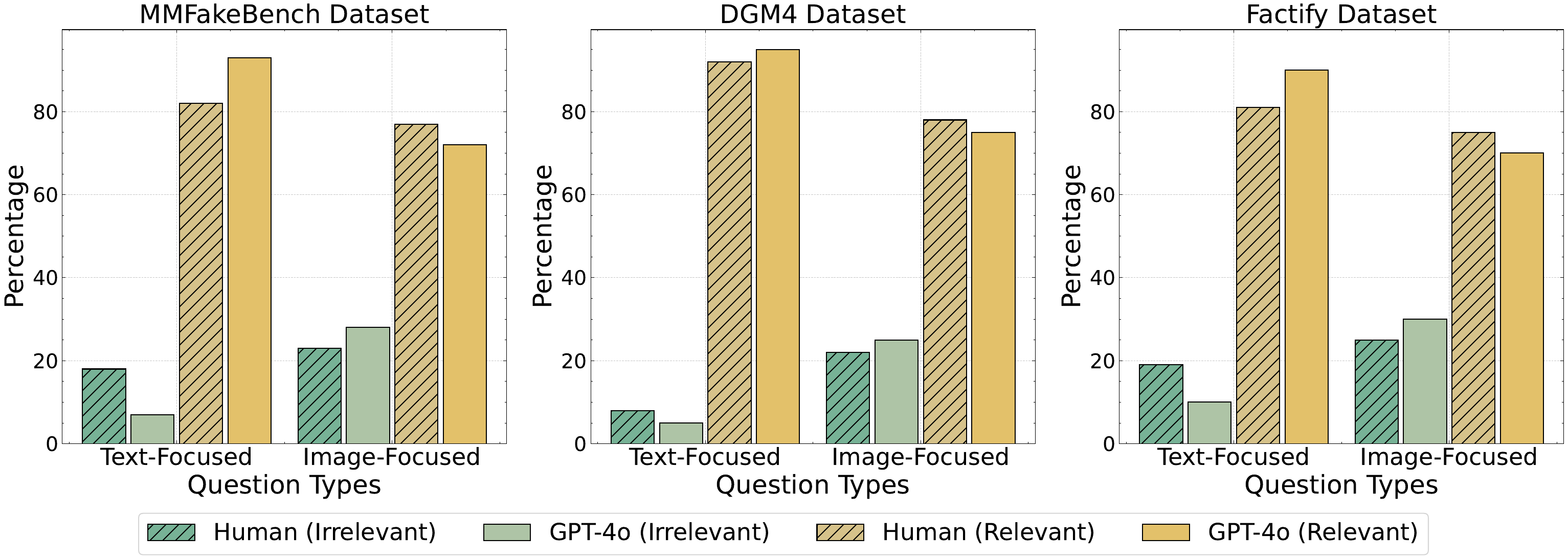}
	\caption{Human and GPT-4o Question Quality Evaluations Across Datasets (50 Samples per Dataset-Modality).}
	\label{fig:50_Samples}
	\vspace{-3mm}
\end{figure*}

\section{Task Definition}
We define the task of multimodal fact-checking as a multiclass classification problem. Given a news article $text_i$ and accompanying image $img_i$ as input,  we aim to classify the news into one of the following categories:
\begin{itemize}
    \item \textbf{Real ($y = 0$):} The news is factually accurate and consistent.  
    \item \textbf{Textual Veracity Distortion (TVD, $y = 1$):} False or misleading claims in the text.
    \item \textbf{Visual Veracity Distortion (VVD, $y = 2$):} Manipulated or misleading images.
    \item \textbf{Cross-Modal Mismatch (CMM, $y = 3$):} Inconsistencies between text and image.
\end{itemize}
\noindent This approach allows for a finer classification of misinformation, improving targeted fact-checking.  


\section{RQ1: Can LLMs Generate Relevant FCQs?}

In this section, we discuss the FCQ generation phase of \textsc{Lrq-Fact}. Specifically, \textsc{Lrq-Fact} produces visual (Sec.~\ref{visual FCQs Generation}) and textual FCQs (Sec.~\ref{textual FCQs Generation}). We then evaluate the quality of the generated FCQs using a combination of LLM-based and human assessments (Sec.~\ref{FCQ Quality Evaluation}).

\subsection{Visual FCQs Generation}\label{visual FCQs Generation}
In the first stage, \textsc{Lrq-Fact} formulates targeted visual FCQs based on the news article to verify whether the visual content aligns with claims in the news article. Importantly, the LLM does not have direct access to the image itself; rather, it generates questions based solely on the article’s content, anticipating what aspects might be depicted in an accompanying image. To guide this process, we employ a structured prompt (see Figure~\ref{fig:prompt_img_q} in Appendix~\ref{prompts}) that instruct the LLM to focus on objects, settings, interactions, and potential manipulations, to enable a comprehensive verification of accuracy, consistency, and authenticity.

For instance, given a news describing a sporting event, the LLM may generate questions such as:
\begin{itemize}
    \item \textit{What sport is being played in the image?}
    \item \textit{Is the pitcher actively throwing the ball?}
    \item \textit{Are there visible spectators?}
    \item \textit{Is the image AI-generated or a real photograph?}
\end{itemize}

\subsection{Textual FCQs Generation}\label{textual FCQs Generation}
The second stage generates fact-checking questions that examine the factual accuracy of textual claims, including dates, names, and events, mimicking human verification methods. Leveraging in-context learning, the LLM formulates precise queries to validate these claims, guided by a structured prompt that ensures relevance and depth in question generation (see Figure~\ref{fig:prompt_text_q} in Appendix~\ref{prompts}). For example, if an article states that an umpire performed a ``jump action'' during a pitch, the LLM might ask:

\begin{itemize}
    \item \textit{Is it common for an umpire to jump during a pitch in baseball?}
    \item \textit{Does “launching the ball” align with standard baseball terminology?}
    \item \textit{Are there baseball rules requiring an umpire to jump while a pitch is thrown?}
\end{itemize}

\subsection{FCQ Quality Evaluation}
\label{FCQ Quality Evaluation}
To ensure a cost-efficient evaluation process, we use human annotators and LLM-as-a-judge to assess the relevance of the generated FCQs. We first establish structured evaluation criteria by incorporating best practices from leading fact-checking organizations~\cite{snopes, PolitiFact, factcheck}.
More details can be found in Table~\ref{tab:tabel_criteria} in Appendix~\ref{FCQ_Quality}. Specifically, we employ ten predefined criteria to guide the LLM judge in FCQ quality evaluation. These criteria cover aspects such as {\it critical thinking, analytical depth, precision, factual accuracy, logical consistency, and source credibility} (see Figure~\ref{fig:prompt_question_quality} in Appendix~\ref{prompts}).

To validate the reliability of the LLM judge, we compare the results of LLM and human annotators (Figure~\ref{fig:50_Samples}) using Fleiss' Kappa correlation~\cite{fleiss1971measuring}. In particular, we randomly sample 50 instances from each dataset, with each instance containing five visual and five textual FCQs. Two human annotators evaluate these FCQs to decide whether they satisfy all predefined criteria.

\begin{figure}[H]
	\centering
	\includegraphics[width=1\columnwidth]{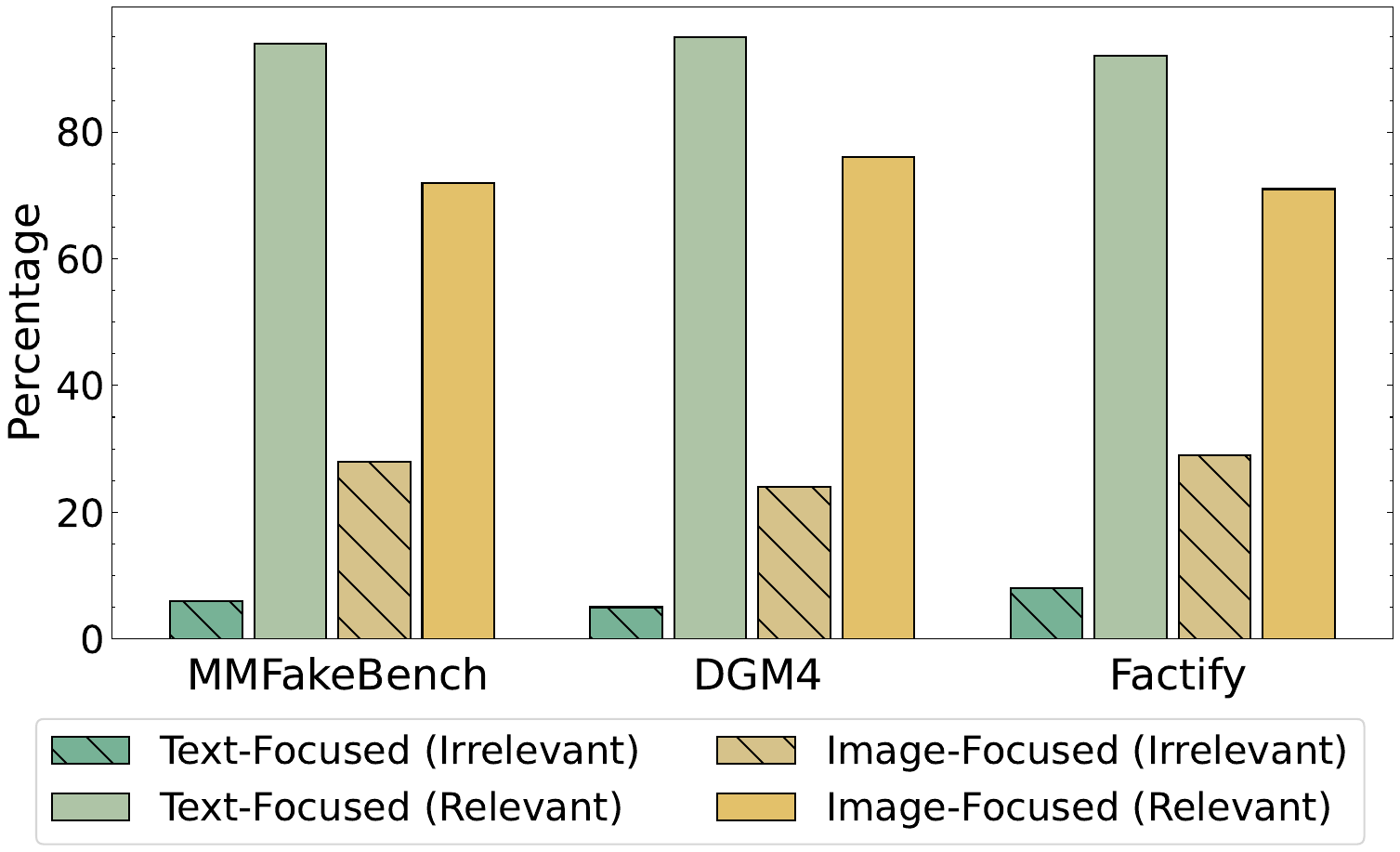}
	\caption{GPT-4o Evaluation of Question Relevance Across Datasets (1000 Samples per Dataset-Modality).}
	\label{fig:1000_Samples}
	\vspace{-2mm}
\end{figure}

\begin{table}[h]
\small
\centering
\setlength{\extrarowheight}{0.1cm}
\begin{tabular}{lccc}
\toprule[1.5pt]
\diag{.2em}{1.4cm}{\tiny\textbf{{Questions}}}{\tiny{\textbf{Dataset}}} & \textbf{MMFakeBench} & \textbf{DGM4} & \textbf{Factify} \\
\hline
Textual & 0.78 & 0.83 & 0.80 \\
Visual & 0.79 & 0.82 & 0.81 \\
\toprule[1.5pt]
\end{tabular}
\caption{Fleiss' Kappa score between human and LLM evaluations of question relevancy.}
\vspace{-2mm}
\label{tab:correlation_scores_evaluation}
\end{table}
Our findings show that LLM-generated FCQs are highly relevant (Table~\ref{tab:correlation_scores_evaluation}), with substantial agreement between human annotators and LLM judge. The high Fleiss’ Kappa scores show the potential to extend the LLM-based quality evaluation to all FCQs. As shown in Figure~\ref{fig:1000_Samples}, nearly $73\%$ of visual and $93.6\%$ of textual FCQs are relevant. These results highlight the effectiveness of LLMs in generating high-quality FCQs.

\section{RQ2: Can LLM-Generated FCQs improve Multimodal Fact-Checking?}
In this section, we first present the remaining components of \textsc{Lrq-Fact}, which includes answering the generated FCQs (Sec.~\ref{Image Description Generation}, \ref{Answering visual FCQs via VLM}, and \ref{rag_section}) and a Rule-Based Decision-Maker (Sec.~\ref{Rule-Based Decision-Maker}). We then investigate RQ2 with extensive experiments and analysis on three datasets. We also provide a case study to showcase the effectiveness of FCQs (Sec.~\ref{Case Study}).

\subsection{Image Description Generation}\label{Image Description Generation}

Since our framework performs the fact-checking in the textual space, we first generate a detailed textual description of the image. The aim is to identify the scene/content/action in the image so that the rule-based stage of the pipeline can use this information to check for consistency between information depicted in the news text vs. shown in the news image. To achieve this, we prompt (see Figure~\ref{fig:prompt_img_description} in Appendix~\ref{prompts}) the VLM to generate a summary of the image, instructing it to ensure it to capture key elements of the scene depicted. Formally, given the image $img_i$, we use the VLM to generate its corresponding description $des_i$:
\begin{equation}
    des_i = {\rm VLM}(img_i).
\end{equation}

\begin{figure}[t]
	\centering
	\includegraphics[width=1\columnwidth]{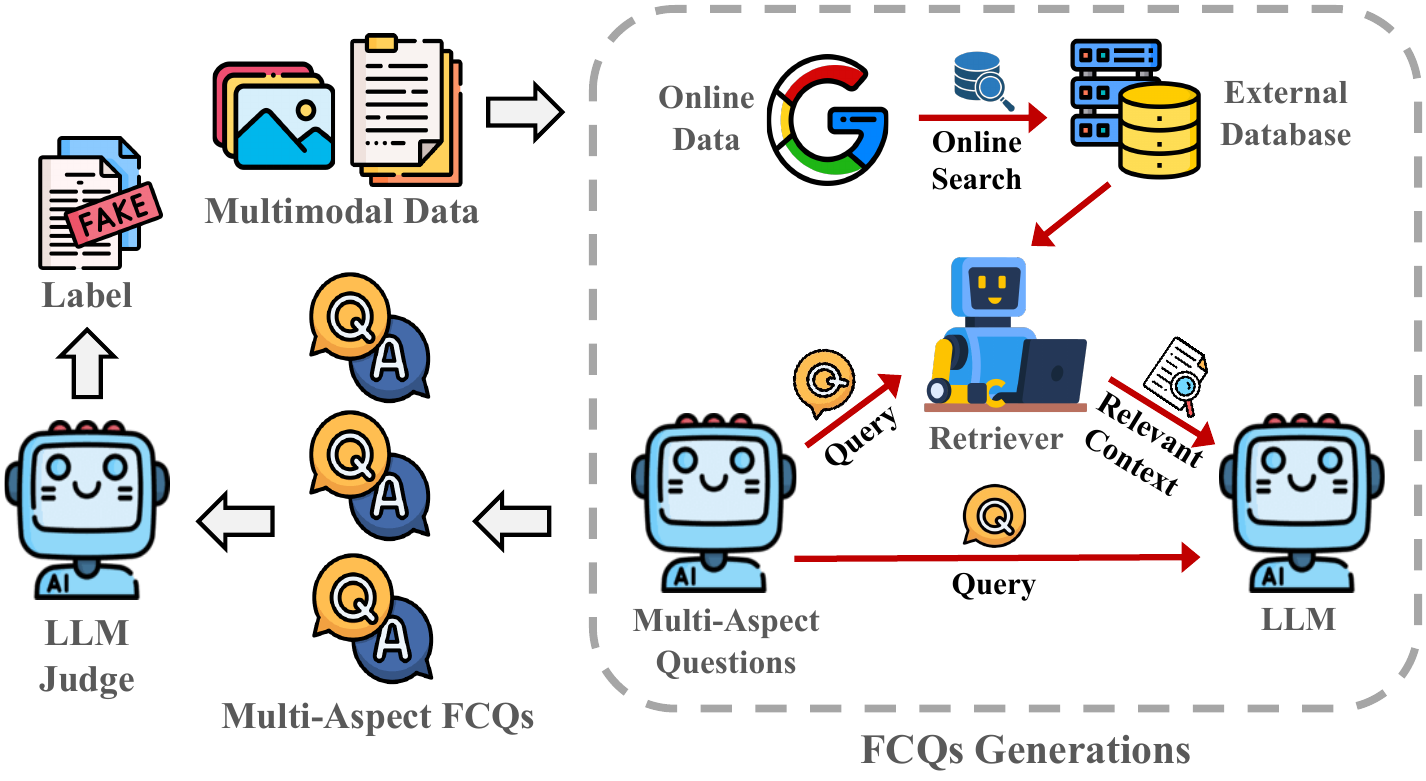}
	\caption{The proposed framework, \textsc{Lrq-Fact}, draws on insights from human fact-checking process.}
	\label{fig:pipeline}
	\vspace{-5mm}
\end{figure}

\subsection{Answering Visual FCQs via VLM}\label{Answering visual FCQs via VLM}

To answer the Visual FCQs, a VLM extracts relevant visual details, allowing for a clear evaluation of how well the image matches the text:
\begin{equation}
    {ques_i^{v_1},...ques_i^{v_m}} = {\rm LLM}(text_i),
\end{equation}
\begin{equation}
    ans_i^{v_j} = {\rm VLM}(img_i, ques_i^{v_j}), \; 0<j<m.
\end{equation}

\subsection{Answering Textual FCQs via RAG}\label{rag_section}
While LLMs are powerful in language generation, they are prone to hallucinations, leading to inaccurate or unverified information. To enhance factual reliability, we employ Retrieval-Augmented Generation (RAG), which grounds responses in external, verifiable sources. This is especially important for newly emerging topics in news articles since the LLM may not have knowledge of such topics due to its knowledge cutoff. We implement an automated online search by using the Google Web Search
API~\cite{google_web_apis} to gather relevant news articles for each claim. We identify the top 10 sources, extract their textual content, and compile a factual document containing relevant information. Next, we use LangChain~\cite{Chase_LangChain_2022} to retrieve the most relevant passages from this document for fact-checking questions, ensuring that only contextually accurate information is used for verification.
Finally, an LLM generates answers based on the retrieved content, reducing hallucinations and improving factual accuracy. To guide the LLM effectively, we employ a carefully designed prompt (see Figure~\ref{fig:prompt_text_a_RAG} in Appendix~\ref{prompts}) that instructs the model to rely solely on the provided factual evidence. This process is formulated as follows:
\vspace{-3mm}
\begin{equation}
    Doc_i = {\rm Search Online} (text_i)
\end{equation}
\begin{equation}
    \begin{aligned}
        RelContent_i &= {\rm Retriever}(Doc_i, ques_i^{t_j})\\
        ans_i^{t_j} &= {\rm LLM}(RelContent_i, ques_i^{t_j})\\
        &\quad\quad\quad 0 < j < n.
    \end{aligned}
\end{equation}
\vspace{-7mm}

\subsection{Rule-Based Decision-Maker}\label{Rule-Based Decision-Maker}
To enhance the judge LLM's ability to effectively utilize multimodal evidence and make accurate predictions, we introduce a rule-based decision-maker module. This module guides the judge LLM to follow expert-like reasoning steps:

\begin{itemize}
    \item \textbf{General Instructions:} We provide guidelines to help judge LLMs establish connections between QA analyses and specific labels. For example, if evidence from FCQs contradicts a claim, the decision-maker assigns it the label ``TVD''.
    \item \textbf{Additional Guidelines:} We outline supplementary instructions, such as analyzing facial expressions, identifying unrealistic elements, and verifying cross-modal consistency, enabling judge LLMs to conduct a human-like examination.
    \item \textbf{Output Format:} We specify the required output format, including the judgment, a fine-grained misinformation label, and a detailed explanation.
\end{itemize}

For each piece of analyzed content, the decision-maker provides: (1) a final judgment $j_i \in $\{Real, Fake\}, (2) a label $l_i \in$ \{Textual Veracity Distortion, Visual Veracity Distortion, Mismatch\} identifying the specific issue, and (3) a detailed explanation of the decision-making process $e_i$. This rule-based approach ensures that the framework provides clear, evidence-based conclusions, carefully weighing the alignment between textual and visual information to detect misinformation. This process can be formulated as:
\begin{equation}
    qa_i^{v} = \oplus_{j=1}^{m}[ques_i^{v_j},ans_i^{v_j}],
\end{equation}
\begin{equation}
    qa_i^{t} = \oplus_{j=1}^{n}[ques_i^{t_j},ans_i^{t_j}],
\end{equation}
\begin{equation}
    j_i, l_i, e_i = {\rm LLM}(text_i, img_i, des_i, qa_i^{v}, qa_i^{t}), 
\end{equation}
where $\oplus$ denotes the concatenation operation, and $m$ and $n$ represent the number of visual and textual FCQs, respectively.

\subsection{Datasets}\label{Datasets}
\noindent\textbf{MMFakeBench}~\cite{liu2024mmfakebench} contains 11,000 image-text pairs. It goes beyond the assumption of single-source forgery and presents samples with \textit{``Real''}, \textit{``Textual Veracity Distortion''}, \textit{``Visual Veracity Distortion''}, and \textit{``Cross-modal Consistency Distortion''}, with both human- and machine-generated images.

\noindent\textbf{DGM$^4$}~\cite{shao2023detecting} is a large-scale multimodal dataset comprising 230,000 image-text paired samples. Image manipulation in the dataset involves \textit{``face swapping and facial emotion editing''}, while text manipulation includes \textit{``sentence replacement and textual sentiment editing''}. The DGM$^4$ dataset is constructed based on the VisualNews dataset \cite{liu2020visual}, which collects data from multiple news agencies.

\noindent\textbf{Factify}~\cite{mishra2022factify} is a multimodal fact-checking benchmark comprising 50,000 data points, each consisting of a textual claim, an associated image, and corresponding reference documents. It is categorized into three main classes: \textit{``Support''}, \textit{``NotEnoughInfo''}, and \textit{``Refute''}, with finer-grained labels for detailed evaluation.

More details about the datasets are provided in Appendix~\ref{dataset}.

\begin{table*}[t]
\centering
\resizebox{\textwidth}{!}{%
\begin{tabular}{l c c c c c c c}

\toprule[1.5pt]
\multirow{2}{*}{\textbf{Backbone}} & \multirow{2}{*}{\textbf{Approach}} & \multicolumn{2}{c}{\textbf{MMFakeBench}} & \multicolumn{2}{c}{\textbf{DGM4}} & \multicolumn{2}{c}{\textbf{Factify}} \\ \cmidrule(lr){3-4} \cmidrule(lr){5-6} \cmidrule(lr){7-8}
& & \textbf{F1$\uparrow$}        & \textbf{ACC$\uparrow$}        & \textbf{F1$\uparrow$}     & \textbf{ACC$\uparrow$}    & \textbf{F1$\uparrow$}      & \textbf{ACC$\uparrow$}      \\ \midrule

VILA {\cite{lin2024vila}} & SP & 11.5 & 30.0 & 19.4 & 19.8 & 25.6 & 27.7\\ 
InstructBLIP {\cite{dai2023instructblipgeneralpurposevisionlanguagemodels}} & SP & 13.7 & 28.8 & 19.2 & 19.8 & 24.3 & 26.9\\ 
BLIP-2 {\cite{li2023blip}} & SP & 16.7 & 32.8 & 18.1 & 19.1 & 26.6 & 28.6\\  
LLaVA-1.6 {\cite{liu2024llava}} & SP & 25.7 & 40.4 & 32.5 & 39.4 & 51.3 & 56.2 \\ 
\midrule
GPT-4V-1.7T {\cite{openai20234v}} & SP & 51.0 & 54.0 & 42.3 & 51.5 & 64.2 & 68.2\\
GPT-4o {\cite{openai20234v}} & SP & 49.2 & 60.9 & 39.9 & 55.9 & \underline{72.5} & \underline{71.2}\\ \hline 
\textsc{Lrq-Fact} (w/o RAG) & FCQs & \underline{66.5} & \underline{65.5} & \underline{45.8} & \underline{58.0} & - & -\\ 
\textsc{Lrq-Fact} (w/ RAG) & FCQs & \textbf{71.6} & \textbf{70.8} & \textbf{49.2} & \textbf{62.3} & \textbf{75.2} & \textbf{73.1}\\ 
\toprule[1.5pt]
\end{tabular}%
}
\caption{The last two rows represent the results of \textsc{Lrq-Fact}. Standard prompt (SP) refers to a generic prompt without fact-checking questions. \textbf{Bold}: best result; \underline{Underline}: second best result.}
\label{tab:comparison-models-mclass}
\vspace{-5mm}
\end{table*}

\subsection{Experiment Details and Settings}\label{Experiment Details and Settings}
For MMFakeBench and DGM$^4$, we sample 1,000 validation instances, ensuring a balanced distribution: 300 \textit{Real}, 300 \textit{TVD}, 100 \textit{VVD}, and 300 \textit{CMM}. DGM$^4$ labels are mapped as follows: \textit{real} (``orig''), \textit{VVD} (``face swap,'' ``face attribute''), \textit{TVD} (``text attribute''), and \textit{mismatch} (``text swap'') to ensure consistency across datasets. For Factify, we sample 750 validation instances: 300 Support (\textit{Support Multimodal}, \textit{Support Text}), 300 NotEnoughInfo ( \textit{Insufficient Text}, \textit{Insufficient Multimodal}), and 150 Refute, maintaining a balanced evaluation.  

\textsc{Lrq-Fact} integrates both LLMs and VLMs to handle the tasks of question generation and answering. For generating visual and textual questions, we use the GPT-4o \cite{achiam2023gpt}, along with Paligemma~\cite{beyer2024paligemma} and LLaMA 3.1~\cite{vavekanand2024llama} as different backbones for VLMs and LLMs respectively. For each modality, we generate five questions by employing in-context learning, where 10 example questions are provided for each modality to guide the model in generating high-quality and relevant questions. To answer fact-checking questions, we employ two settings: (1) using only LLM knowledge and (2) a RAG approach, where an external knowledge retrieval module fetches relevant supporting documents before generating responses. In the decision-making phase, GPT-4o is also used to leverage the FCQs for fact verification. Known for its strong reasoning and rule-following capabilities, GPT-4o assesses the veracity of the content and offers clear rationales for its conclusions. All experiments are conducted on NVIDIA 40GB V100 GPUs.

\vspace{-3mm}
\subsubsection{Evaluation Metrics}
To assess the performance of the various baselines, we employ a multi-class classification approach. Following the practices established in prior works \cite{zhang2023towards,qian2021counterfactual,chen2023causal}, we use the widely adopted macro-F1 score as our primary evaluation metric. The macro-F1 score provides a balanced measure of precision and recall through their harmonic mean, ensuring fair evaluation across all classes. In addition to the macro-F1 score, we also report macro-accuracy as complementary metrics, offering a more comprehensive understanding of model performance.

\subsubsection{Comparison Models}
We select a diverse range of VLMs as baseline models for comparison. These include: InstructBLIP \cite{dai2023instructblipgeneralpurposevisionlanguagemodels}, VILA \cite{lin2024vila}, BLIP-2 \cite{li2023blip}, LLaVA-1.6 \cite{liu2024llava} and  the closed-source GPT-4V-1.7T, GPT-4o \cite{openai20234v} models. 

\subsection{Experiment Analysis}\label{Experiment Analysis}
To assess whether LLM-generated FCQs enhance multimodal fact-checking, we conduct a three-stage analysis: (1) a comparison between our \textsc{Lrq-Fact} and other baseline methods, (2) an ablation study to measure the impact of generated questions on fact-checking performance and (3) an in-depth examination of how different types of questions contribute to improving verification accuracy across modalities.

\begin{table*}[h]
\centering
\resizebox{\textwidth}{!}{
\begin{tabular}{lcccccc}
\toprule[1.5pt]
\multirow{2}{*}{\textbf{Method}}                                                                      & \multicolumn{2}{c}{\textbf{MMFakeBench}} & \multicolumn{2}{c}{\textbf{DGM4}} & \multicolumn{2}{c}{\textbf{Factify}} \\ \cmidrule(lr){2-3} \cmidrule(lr){4-5} \cmidrule(lr){6-7}
& \textbf{F1}        & \textbf{ACC}        & \textbf{F1}     & \textbf{ACC}    & \textbf{F1}      & \textbf{ACC}      \\ \midrule

\textsc{Lrq-Fact} (VLM: Paligemma \& LLM: LLaMA 3.1)                                                                              &                51.2 &     56.4            &      41.2       &    53.6         &      66.2        &        65.3       \\ \midrule

\ \ \ w/ Textual FCQs & 62.5\textsubscript{+22.1\%} & 59.1\textsubscript{+4.8\%} & 43.3\textsubscript{+5.1\%} & 54.1\textsubscript{+0.9\%} & 64.8\textsubscript{-2.1\%} & 60.9\textsubscript{-6.7\%} \\ \addlinespace[2pt] \hdashline[1pt/1pt] \addlinespace[2pt]

\ \ \ w/ Visual FCQs (w/o RAG) & 59.4\textsubscript{+16.0\%} & 57.3\textsubscript{+1.6\%} & 46.1\textsubscript{+11.8\%} & 53.9\textsubscript{+0.5\%} &
- & - \\ \addlinespace[2pt] \hdashline[1pt/1pt] \addlinespace[2pt]

\ \ \ \begin{tabular}[c]{@{}l@{}}w/ Visual \& Textual FCQs (w/o RAG)\end{tabular} & 62.8\textsubscript{+22.6\%}               & 61\textsubscript{+8.1\%}                & 47.7\textsubscript{+15.7\%}            & 54.8\textsubscript{+2.2\%}            & -                & -                 \\ \addlinespace[2pt] \hdashline[1pt/1pt] \addlinespace[2pt]

\ \ \ w/ Textual FCQs (w/ RAG) & 
61.7\textsubscript{+20.5\%} & 63.2\textsubscript{+12.1\%} & 
48.3\textsubscript{+17.2\%} & 61.7\textsubscript{+15.1\%} & 
69.5\textsubscript{+5.0\%} & 67.2\textsubscript{+2.9\%} \\ \addlinespace[2pt] \hdashline[1pt/1pt] \addlinespace[2pt]

\ \ \ \begin{tabular}[c]{@{}l@{}}w/ Visual \& Textual FCQs (w/ RAG)\end{tabular}  & 64.2\textsubscript{+25.3\%}               & 64.8\textsubscript{+14.8\%}                & 48.6\textsubscript{+17.9\%}            & 62.1\textsubscript{+15.8\%}            & 73.2\textsubscript{+10.5\%}             & 71.5\textsubscript{+9.4\%}              \\ 

\midrule
\textsc{Lrq-Fact} (VLM \& LLM: GPT-4o)                                                                             & 49.2               & 60.9                & 39.9            & 55.9            & 72.5             & 71.2              \\ \midrule
\ \ \ w/ Textual FCQs                                                                        & 64.3\textsubscript{+30.7\%}               & 62.1\textsubscript{+2.0\%}                & 40.4\textsubscript{+1.3\%}            & 54.3\textsubscript{-2.9\%}            & 71.1\textsubscript{-1.9\%}             & 68.7\textsubscript{-3.5\%}              \\ \addlinespace[2pt] \hdashline[1pt/1pt] \addlinespace[2pt]
\ \ \ w/ Visual FCQs (w/o RAG)                                                          & 59.6\textsubscript{+21.1\%}               & 61.7\textsubscript{+1.3\%}                & 47.2\textsubscript{+18.3\%}            & 59.7\textsubscript{+6.8\%}            & -                & -                 \\ \addlinespace[2pt] \hdashline[1pt/1pt] \addlinespace[2pt]
\ \ \ \begin{tabular}[c]{@{}l@{}}w/ Visual \& Textual FCQs (w/o RAG)\end{tabular} & \underline{66.5}\textsubscript{+35.2\%}               & \underline{65.5}\textsubscript{+7.5\%}                & 45.8\textsubscript{+14.8\%}            & 58.0\textsubscript{+3.8\%}            & -                & -                 \\ \addlinespace[2pt] \hdashline[1pt/1pt] \addlinespace[2pt]
\ \ \ w/ Textual FCQs (w/ RAG)                                                           & 61.8\textsubscript{+25.2\%}               & 64.7\textsubscript{+6.2\%}                & \textbf{49.4}\textsubscript{+23.8\%}            & \textbf{62.5}\textsubscript{+11.8\%}            & \underline{74.0}\textsubscript{+2.1\%}             & \underline{72.5}\textsubscript{+1.8\%}              \\ \addlinespace[2pt] \hdashline[1pt/1pt] \addlinespace[2pt]
\ \ \ \begin{tabular}[c]{@{}l@{}}w/ Visual \& Textual FCQs (w/ RAG)\end{tabular}  & \textbf{71.6}\textsubscript{+45.5\%}               & \textbf{70.8}\textsubscript{+16.3\%}                & \underline{49.2}\textsubscript{+23.3\%}            & \underline{62.3}\textsubscript{+11.5\%}            & \textbf{75.2}\textsubscript{+3.7\%}             & \textbf{73.1}\textsubscript{+2.7\%}              \\ \toprule[1.5pt] 
\end{tabular}
}
\caption{Ablation study result. The subscript values indicate the percentage improvement over the GPT-4o baseline.}
\vspace{-4mm}
\label{tab:ablation_study}
\end{table*}

\subsubsection{Comparison with Other Methods}
Table~\ref{tab:comparison-models-mclass} compares our \textsc{Lrq-Fact} model with baseline methods. Our proposed approach consistently outperforms all baselines across various datasets, regardless of whether external knowledge is introduced (w/ RAG) or not (w/o RAG). This highlights the effectiveness of incorporating FCQs in enhancing fact-checking performance.

\subsubsection{Effect of LLM-Generated FCQs on Fact-Checking Performance}
Table~\ref{tab:ablation_study} reports F1 scores and accuracy (ACC) for various ablation configurations, comparing the baseline GPT-4o model with visual and textual FCQs variations.

\noindent\textbf{Overall Improvements:} Incorporating FCQs enhances fact-checking performance across all datasets. For MMFakeBench, the F1 score improves from 49.2 to 71.6 (+45.5\%), and accuracy rises from 60.9 to 70.8 (+16.3\%). Similarly, for DGM$^4$, the highest F1 score increases from 39.9 to 49.2 (+23.3\%), while accuracy improves from 55.9 to 62.3 (+11.5\%). Factify also benefits from FCQs integration, with F1 increasing from 72.5 to 75.2 (+3.7\%) and accuracy from 71.2 to 73.1 (+2.7\%).  

\noindent\textbf{Impact of Visual FCQs:} Visual FCQs alone yield noticeable improvements in both F1 and accuracy across all datasets. For MMFakeBench, F1 improves by 30.7\% (from 49.2 to 64.3), while accuracy rises slightly (+2.0\%). A similar trend is observed for DGM$^4$, where F1 improves by 1.3\% and accuracy by 2.9\%. This improvement can be attributed to the fact that visual FCQs encourage the model to focus more on visual details, prompting a deeper analysis of the image content.

\noindent\textbf{Impact of Textual FCQs without RAG:} The addition of textual FCQs without explicit external evidence shows significant gains, especially in DGM$^4$, where the F1 score rises from 39.9 to 47.2 (+18.3\%), and accuracy improves by 6.8\%. However, this setting is not evaluated for Factify because the task in this dataset is verification rather than detection, and Factify already provides factual reference documents for verification. This improvement can be attributed to the fact that textual FCQs systematically capture all claims within the news and probe their validity. By generating targeted questions, the model breaks down the text into specific factual assertions, allowing for a more structured verification process. This focused approach helps detect inconsistencies, misinterpretations, or misleading statements within the text, ultimately leading to more accurate fact-checking.  

\noindent\textbf{Impact of Textual FCQs with RAG:} When textual FCQs incorporate external evidence, fact-checking performance further improves. For MMFakeBench, F1 increases to 61.8 (+25.2\%), and accuracy improves by 6.2\%. In DGM$^4$, this approach achieves the highest F1 (49.4, +23.8\%) and accuracy (62.5, +11.8\%), highlighting the importance of retrieved evidence. Factify also benefits, with F1 rising to 74.0 (+2.1\%) and accuracy to 72.5 (+1.8\%).

These results confirm that integrating LLM-generated FCQs, particularly when paired with external evidence, significantly enhances multimodal fact-checking performance.

\subsubsection{Effect of Fact-Checking Questions on Fact-Checking Performance}  
\begin{figure*}[h]
	\centering
	\includegraphics[width=0.9\textwidth]{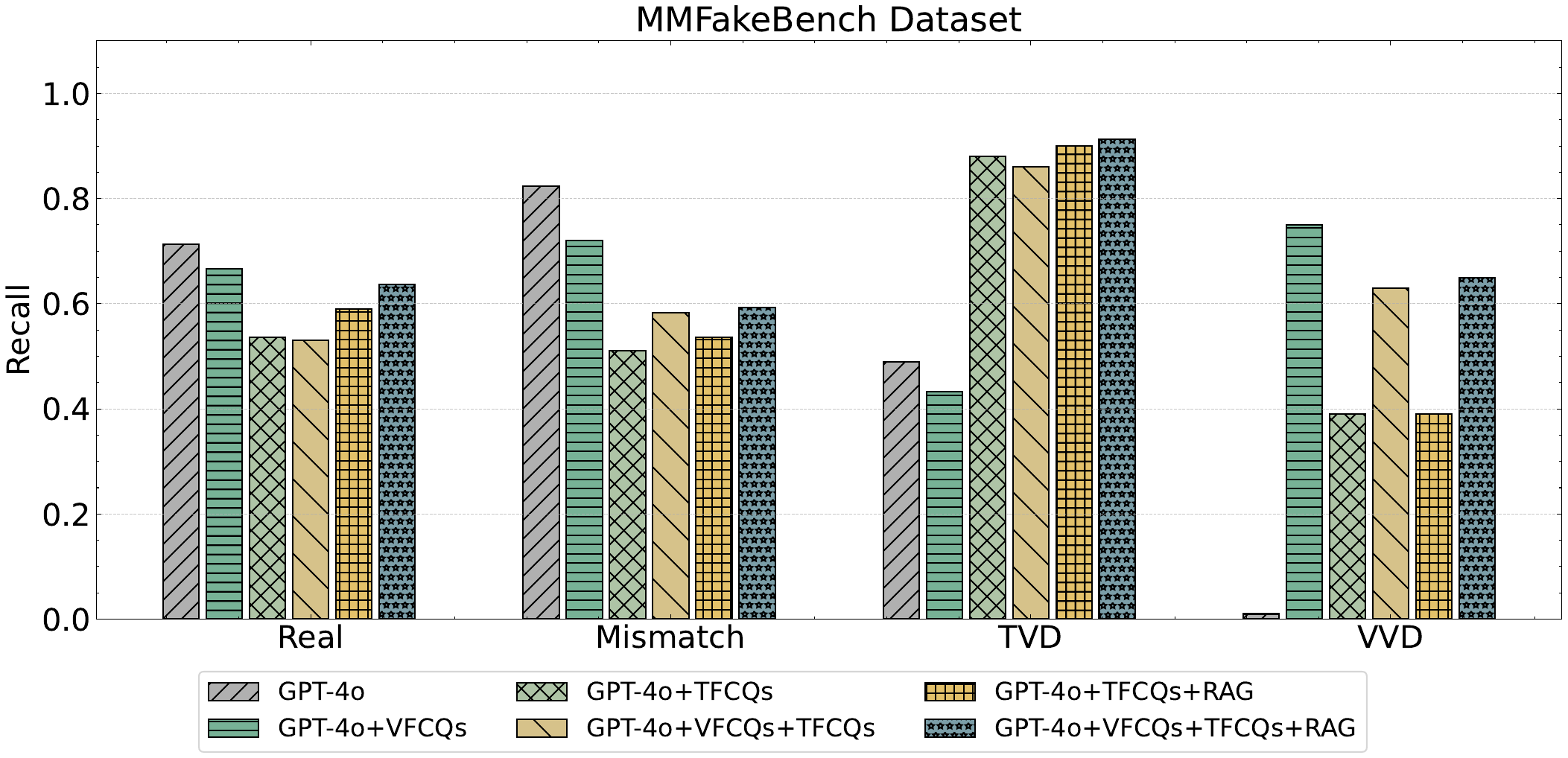}
	\caption{Detailed ablation study result. TFCQs and VFCQs represent Textual FCQs, Visual FCQs respectively.}
	\label{fig:detailed_ablation_study}
    \vspace{-4mm}
\end{figure*}

To further investigate how different types FCQs impact verification performance, we analyze recall scores across different manipulation categories within the MMFakeBench dataset. Figure~\ref{fig:detailed_ablation_study} presents recall values for various settings, including visual FCQs (VFCQs), textual FCQs (TFCQs), and combinations of both with and without RAG.  

\noindent\textbf{Impact on Real and Mismatch Cases:} While FCQs improve fact-checking performance overall, recall scores for the \textit{Real} and \textit{Mismatch} categories decrease. This phenomenon occurs because, in the absence of structured questioning, the model does not apply strict fact-checking criteria and tends to classify most news samples as either \textit{Real} or \textit{Mismatch}. However, with FCQs, the model adopts a more expert-like approach, becoming more cautious before confirming a claim as real. This increased scrutiny aligns with human verification processes, reducing false positives but leading to a lower recall in these categories.  

\noindent\textbf{Impact on Textual Veracity Distortion (TVD):} FCQs significantly enhance recall for cases involving textual manipulation. Without structured questioning, the model relies primarily on its internal knowledge, which may not always be up-to-date or factually accurate. However, incorporating textual FCQs (TFCQs) allows the model to engage in a more structured verification process by utilizing factual information instead of relying solely on pre-trained LLM knowledge. Additionally, integrating external evidence (TFCQs+RAG) enhances the model’s ability to detect inconsistencies in manipulated text, improving misinformation identification. This highlights the value of supplementing LLM-generated responses with retrieved factual data for more reliable fact-checking.

\noindent\textbf{Impact on Visual Veracity Distortion (VVD):} Visually manipulated claims pose a challenge for the baseline model, which often lacks the ability to rigorously assess image alterations. However, incorporating visual FCQs (VFCQs) substantially improves recall by prompting the model to analyze visual content more critically. The highest recall is achieved when both VFCQs and TFCQs are used together with external evidence, reinforcing the importance of cross-modal verification.  

Across all manipulation types, the most effective setup involves combining VFCQs and TFCQs with external evidence (VFCQs+TFCQs+RAG). While this structured questioning approach makes the model more cautious in verifying real claims, it significantly enhances misinformation detection, ensuring a more reliable fact-checking process. These findings highlight how FCQs improve fact-checking by encouraging a more rigorous verification process. By systematically questioning claims across both textual and visual modalities and incorporating factual retrieval mechanisms, FCQs help the model adopt a more expert-like approach, leading to more precise and reliable fact verification.  

\subsection{Case Study}\label{Case Study}
To demonstrate the effectiveness of FCQs in detecting and classifying multimodal misinformation, we provide examples in the Appendix (see Figures~\ref{fig:example1}, \ref{fig:example4}, \ref{fig:example3}, and \ref{fig:example2}) showcasing their impact.

\section{Conclusion}
In this work, we investigate whether LLMs can generate relevant FCQs and whether these LLM-generated FCQs can improve multimodal fact-checking. Through the proposed framework, \textsc{Lrq-Fact}, we demonstrate that LLMs are indeed capable of generating highly relevant and targeted FCQs, effectively addressing a key limitation in AFC systems. Furthermore, our experiments show that incorporating relevant FCQs into the fact-checking process significantly enhances evidence retrieval and improves the overall factuality verification performance. \textsc{Lrq-Fact} outperforms baseline methods, showcasing the effectiveness of FCQ generation in strengthening multimodal fact-checking. These results highlight the potential of LLM-based FCQ formulation as a promising direction in future AFC research, facilitating more reliable and scalable methods.

\section*{Limitations}

While our study demonstrates that LLM-generated FCQs enhance multimodal fact-checking, several limitations remain. One major limitation is the lack of expert-level validation. Although we evaluate FCQ quality using benchmark datasets and human annotations, we do not assess whether the generated questions exhibit reasoning comparable to domain experts. Incorporating expert evaluations, particularly in specialized fields such as medicine or law, could provide a more rigorous assessment of FCQ quality and alignment with high-quality fact-checking standards. 

Another limitation is the absence of a random question baseline. Our experiments compare \textsc{Lrq-Fact} against strong fact-checking baselines, but we do not explicitly test whether randomly generated questions could serve as a control. Introducing a random baseline would help isolate the actual contribution of meaningful FCQ generation from the broader effect of question-driven retrieval. Additionally, our approach is inherently dependent on the capabilities of the underlying LLM. If the model produces vague, misleading, or hallucinated questions, it could negatively impact fact-checking performance by retrieving irrelevant or incorrect evidence. Further investigation into fine-tuned models or more controlled question-generation strategies could mitigate these risks.

Our evaluation process also introduces certain limitations. The FCQs were assessed by PhD students who received detailed instructions and predefined criteria to evaluate question relevance. While this ensures a structured and consistent evaluation process, the annotator pool is relatively small and may not fully represent diverse perspectives. The background knowledge of annotators could influence their judgments, and a broader demographic, including professional fact-checkers or domain experts, may provide a more comprehensive evaluation of FCQ effectiveness.

The reliance on external search engines for evidence retrieval introduces another source of variability. The effectiveness of our approach depends on search engine algorithms, indexing policies, and the availability of high-quality sources, which may not always be consistent. This issue is particularly relevant for fact-checking emerging claims or topics where authoritative sources are scarce. Furthermore, while \textsc{Lrq-Fact} integrates textual and visual evidence for multimodal fact-checking, its reliance on image captions and textual representations of visual content may introduce errors. Exploring direct visual analysis through vision-language models or image embeddings could improve robustness in cases where textual descriptions are insufficient.

Generalization across fact-checking domains remains another open challenge. Although our approach performs well on benchmark datasets, its effectiveness across diverse domains such as political misinformation, scientific fact-checking, and real-time verification is not fully explored. Future work should investigate domain-specific FCQ generation techniques to adapt the framework for specialized fact-checking tasks. Additionally, the computational overhead of our approach may limit its practical deployment. Multiple inference steps, including FCQ generation, evidence retrieval, and multimodal reasoning, contribute to significant resource consumption, making real-time fact-checking more challenging. Optimizing the framework for efficiency would be necessary for large-scale deployment.

Finally, our evaluation process relies on both human annotators and LLM-based scoring to assess FCQ quality. While high agreement between human and model-based evaluations suggests reliability, potential biases in human annotation criteria and systematic artifacts in LLM-generated scoring may still influence results. Future research should explore alternative evaluation metrics and methodologies to ensure robustness and fairness in assessing FCQ effectiveness.

\section*{Ethical Statement}

This work explores the use of LLMs to generate fact-checking questions (FCQs) for automated verification. While our approach enhances misinformation detection, it raises ethical considerations. LLM-generated FCQs may reflect biases present in training data, potentially influencing fact-checking outcomes. Although we evaluate FCQ quality with human reviewers and benchmark datasets, further efforts are needed to ensure fairness and neutrality. 

Another concern is the risk of over-reliance on automation. While LLMs can support fact-checking at scale, they should not replace human judgment, particularly in high-stakes domains like politics and healthcare. Our framework is designed as an assistive tool, reinforcing rather than replacing expert oversight. Privacy considerations are also critical, as our approach retrieves publicly available evidence without processing personally identifiable information. We adhere to ethical data usage practices and fair use policies but recognize the need for continuous alignment with evolving privacy standards. 

Finally, transparency and accountability in AI-driven fact-checking remain essential. Black-box decision-making can undermine trust, emphasizing the importance of explainability. By acknowledging these challenges, we advocate for responsible AI deployment, promoting fairness, human oversight, and transparency in automated fact-checking systems.


\bibliography{main}

\appendix

\section{Appendix}
\subsection{Example Analysis}
\begin{figure*}[t]
	\centering
	\includegraphics[width=1\textwidth]{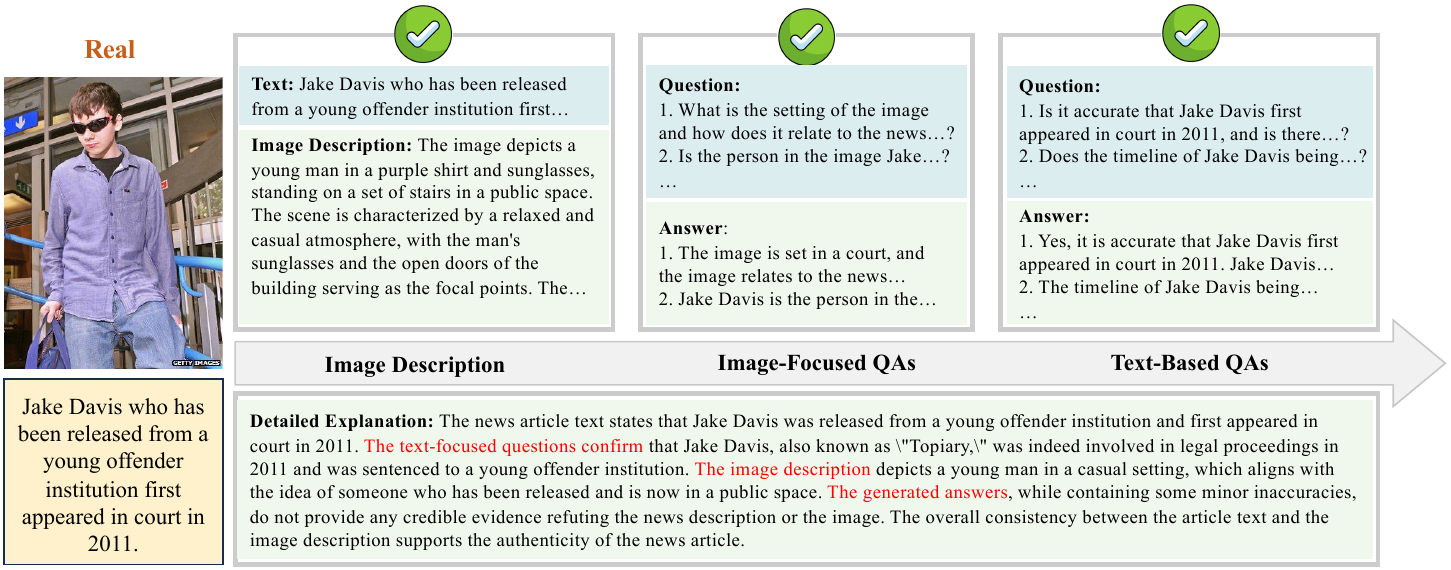}
	\caption{This example case illustrating the alignment between image and text in a fact-checking process. The generated questions verify key elements, ensuring consistency and accuracy in multimodal misinformation detection. This demonstrates how targeted questions and well-constructed descriptions enhance reliable fact-checking outcomes.}
	\label{fig:example1}
\end{figure*}

\noindent\textbf{Real.}
In this case study (Figure \ref{fig:example1}), the description is well-constructed and aligns perfectly with the image and the textual context. The depiction of Jake Davis as a young man in casual clothing, standing in a relaxed manner, accurately reflects the narrative of his release from a young offender institution. The visual context provided in the image adds credibility to the news article, confirming the validity of the description. There are no discrepancies between the image and the text, making the description not only good but also a reliable tool to confirm the factual correctness of the news.

The questions presented in this case are well-formed and reliable. They are designed to extract key details from both the image and the text, ensuring comprehensive verification. The visual questions effectively ask about the setting and identity, which helps in confirming whether the person and location in the image match the article’s claims. The text-based questions aim to validate the timeline and factual details, ensuring a consistent narrative. These questions are precise and structured to get the best possible answers, making them a solid mechanism for cross-verifying facts.

\begin{figure*}[t]
	\centering
	\includegraphics[width=1\textwidth]{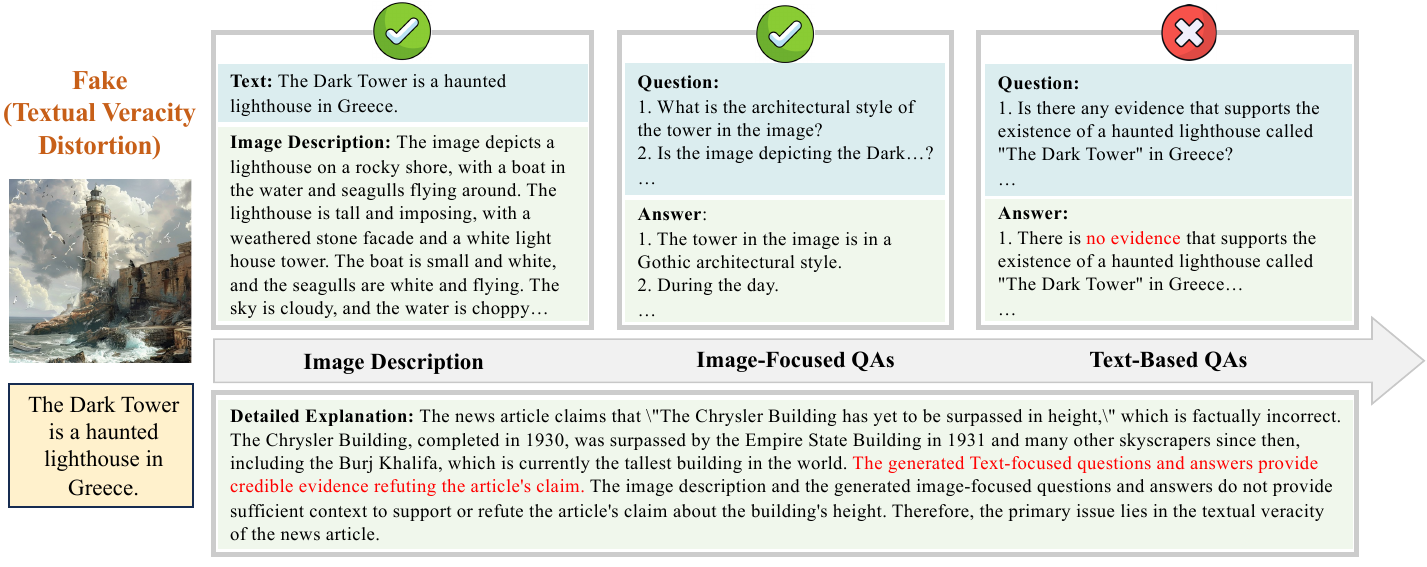}
	\caption{Example case illustrating textual veracity distortion. The image description aligns visually with the content, but fails to support the false textual claim about the haunted lighthouse's location in Greece. The generated questions are designed to detect inconsistencies, providing a thorough framework for fact-checking by scrutinizing both visual and textual elements.}
	\label{fig:example4}
\end{figure*}

\noindent\textbf{Textual Veracity Distortion.}
The description in Figure \ref{fig:example4} accurately depicts a lighthouse in a Gothic architectural style, positioned on a rocky shore with surrounding water and seagulls. The image is valid and corresponds with the article's general theme. However, the description’s alignment with the actual claim in the text—that the lighthouse is haunted and located in Greece—proves to be incorrect. While the description is visually consistent and good, it does not support the erroneous textual claim, showing how important it is to assess both text and visuals in tandem.

The questions in this case are reliable and appropriately structured to identify discrepancies between the image and the text. The visual questions ask about the architectural style and contextual clues from the image, while the text-based questions explore the factual accuracy of the claim that this lighthouse is haunted and located in Greece. The questions provide a good framework for fact-checking by encouraging thorough scrutiny of both visual and textual elements. This ensures that any distortions or misrepresentations in the article are effectively highlighted, making the questions a valuable tool for getting to the truth.\\

\begin{figure*}[t]
	\centering
	\includegraphics[width=1\textwidth]{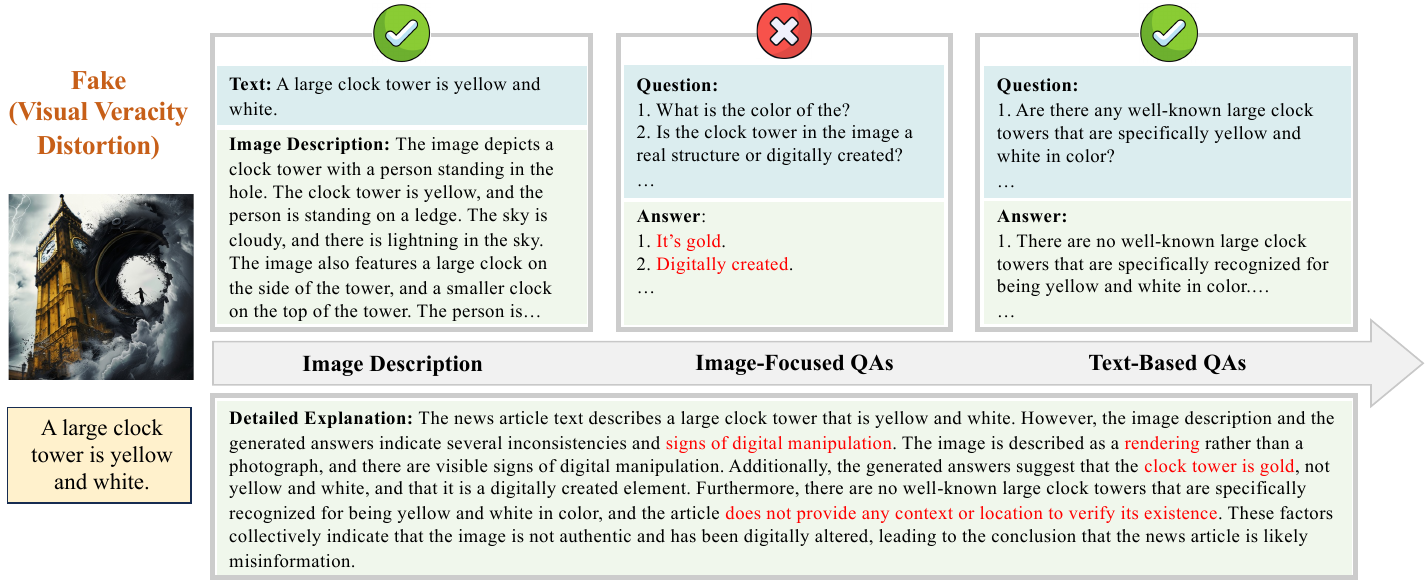}
	\caption{This example case highlighting visual manipulation. The description accurately conveys the textual claim about a yellow and white clock tower, but fails to reflect the digitally altered gold structure seen in the image. The questions focus on detecting visual discrepancies, such as the altered colors, and also probe the existence of such a clock tower, providing a reliable framework for evaluating both the image and text.}
	\label{fig:example3}
\end{figure*}

\noindent\textbf{Visual Veracity Distortion.}
In Figure \ref{fig:example3}, the description of a clock tower in yellow and white is valid and clear, but the image itself shows a structure that is clearly gold and digitally altered. The description is good in terms of clarity and helping readers visualize the article's claim, even though it does not reflect the manipulated nature of the image. This highlights the importance of analyzing the veracity of visuals alongside textual descriptions.

The questions are well-crafted to reveal any visual inconsistencies. The visual questions ask about the color and reality of the clock tower, which are key to identifying that the clock tower has been digitally altered. The text-based questions, which probe the existence of such a clock tower in real life, also help uncover discrepancies. These questions are reliable and precise, aimed at extracting the best possible answers and guiding the evaluation of the article’s claims against the evidence provided by the image.

\begin{figure*}[t]
	\centering
	\includegraphics[width=1\textwidth]{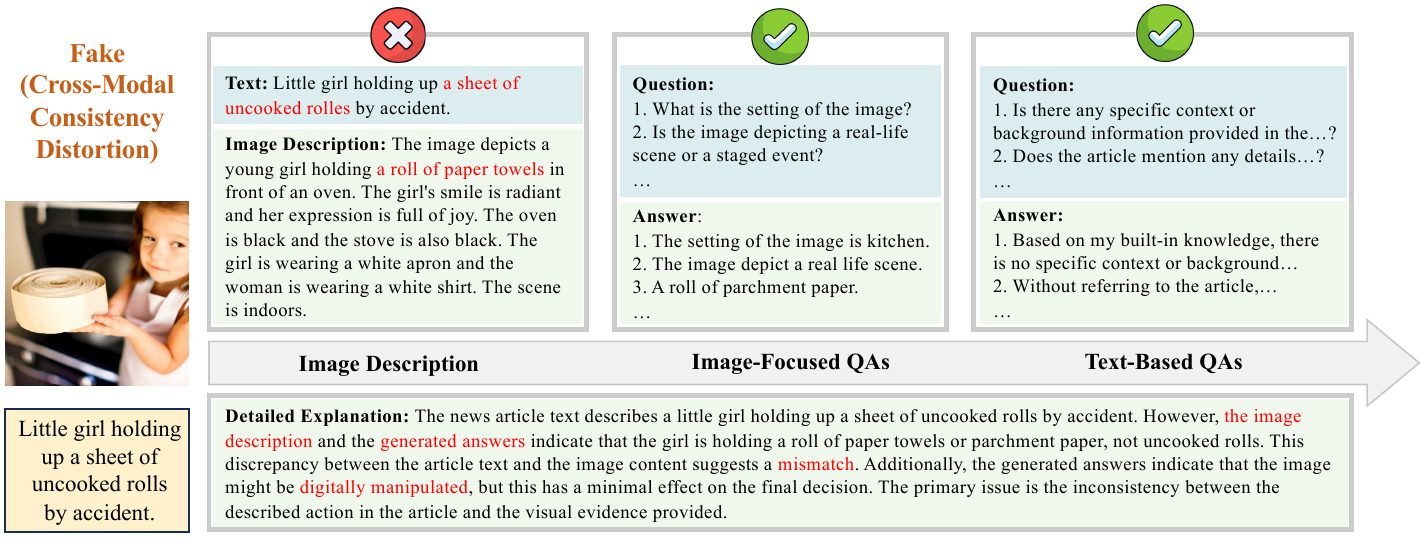}
	\caption{This example case demonstrating cross-modal distortion. The description is clear and helps visualize the article’s scenario of a little girl holding uncooked rolls, while the image actually shows her holding paper towels. This mismatch between text and image points to a cross-modal distortion. The questions are well-crafted to reveal this inconsistency by focusing on both the scene and the object in the girl's hands, providing a reliable framework for identifying the discrepancy between the article and the image. \\
 \ \\
 \ \\
 \ \\
 \ \\}
	\label{fig:example2}
\end{figure*}

\noindent\textbf{Cross-modal Consistency Distortion.}
This case (Figure \ref{fig:example2}) involves a clear mismatch between the text and the image, where the article describes a little girl holding uncooked rolls, while the image shows her holding paper towels. The description is coherent and well-explained, making it a good tool to visualize the scenario presented in the article. However, the inconsistency between the image and the text highlights a cross-modal distortion. Despite this, the description itself remains valid in its own right.

The questions presented are well-designed to highlight the inconsistency between the text and the image. The visual questions ask about the scene and the object the girl is holding, providing clear answers that reveal the mismatch. The text-based questions further confirm this by addressing the article’s lack of accurate description. These questions are well-structured and reliable, allowing for an in-depth examination of both the image and the text to expose cross-modal discrepancies. They guide the analysis toward the best possible answers by focusing on the key elements that need verification.

\subsection{Instruct Prompt for \textsc{Lrq-Fact}}
\label{prompts}
The \textsc{Lrq-Fact} framework employs a series of structured prompts to guide LLMs and VLMs in multimodal fact-checking. These prompts facilitate the generation of detailed image descriptions, contextually relevant questions, and well-informed answers that probe the veracity of both visual and textual content. In the final step, a rule-based decision-maker evaluates the generated questions and answers to provide a final judgment on the consistency between the text and image, ensuring accurate detection of misinformation.

\noindent\textbf{Image Description Prompt.}
The first step is to generate a detailed description of the image, capturing all relevant elements that help assess its consistency with the textual content. This description is crucial for identifying potential inconsistencies or manipulations between the image and the accompanying article. The specific prompt used to generate this description is provided in Figure \ref{fig:prompt_img_description}.

\noindent\textbf{Visual Questions Prompt.}
This stage generates relevant visual questions designed to verify the accuracy, authenticity, and relevance of the visual content in relation to the article. These questions help clarify the image content and assess its relation to the text. The specific prompt for generating these questions is illustrated in Figure \ref{fig:prompt_img_q}.

\noindent\textbf{Visual Answers Prompt.}
After generating the visual questions, this prompt helps in generating answers that analyze the visual content directly from the image. These answers are based on the key elements and actions identified in the image, ensuring that the responses are relevant and insightful. The specific prompt for this is shown in Figure \ref{fig:prompt_img_a}.

\noindent\textbf{Textual Questions Prompt.}
To critically assess the factual claims in the text, this prompt generates relevant questions targeting specific elements such as dates, names, locations, and events. The generated questions aim to challenge the accuracy of the claims made in the article. The specific prompt used for textual questions is shown in Figure \ref{fig:prompt_text_q}.

\noindent\textbf{Textual Answers Prompt.}
After generating textual questions, this prompt enables the model to generate answers using its built-in knowledge. The specific prompt is shown in Figure~\ref{fig:prompt_text_a}. Additionally, we employ a Retrieval-Augmented Generation (RAG) approach to incorporate factual evidence, ensuring more reliable and verifiable responses. These answers help assess factual accuracy and challenge any unsupported claims in the article. The corresponding RAG-based prompt is illustrated in Figure~\ref{fig:prompt_text_a_RAG}.  

\noindent\textbf{Question Quality Assessment Prompt.}  
To evaluate the relevance of generated questions, we use a fact-checking criteria-based prompt that classifies questions as relevant or irrelevant. This assessment considers factors such as alignment with the claim, specificity, and usefulness in verifying factual accuracy. The specific prompt used for this evaluation is illustrated in Figure~\ref{fig:prompt_question_quality}.  

\noindent\textbf{Rule-Based Decision-Maker Prompt.}
After gathering information from the image and text analyses, the rule-based decision-maker evaluates the consistency between modalities and makes a final determination about the article’s veracity. This module provides a detailed explanation for the final judgment. The specific prompt for the rule-based decision-making process is shown in Figure \ref{fig:prompt_judge}.

\subsection{Annotator Details}
To evaluate the quality of LLM-generated FCQs, we recruited two PhD students with backgrounds in NLP and computational linguistics. Annotators were provided with detailed instructions and predefined criteria to assess the relevance of each FCQ to the given claim. The evaluation process aimed to ensure consistency and minimize subjectivity in judgment. While this setup provides structured and knowledgeable assessments, the annotator pool is relatively small and may not fully capture diverse perspectives. Future work could incorporate domain experts or professional fact-checkers to further validate FCQ effectiveness across different fact-checking domains.

\subsection{Criteria for Evaluating FCQ Quality}
\label{FCQ_Quality}

To systematically evaluate the quality of LLM-generated fact-checking questions (FCQs), we developed a structured evaluation framework inspired by best practices from established fact-checking methodologies. Our evaluation process consists of two key components: LLM-based assessment and human evaluation, ensuring a rigorous and reliable analysis of question relevance.

\noindent\textbf{Evaluation Framework}. We designed our evaluation framework to assess the effectiveness of both \textit{visual} and \textit{textual} FCQs. The framework follows ten evaluation criteria, derived from widely accepted fact-checking principles, emphasizing accuracy, credibility, and relevance. These criteria help determine whether the generated FCQs effectively probe factual claims and align with real-world verification standards (Figure \ref{fig:prompt_question_quality}).

One challenge in evaluating FCQs is ensuring question specificity without over-constraining the verification process. A well-formed FCQ should allow multiple valid answers depending on available evidence while still prompting meaningful fact-checking efforts. Additionally, cross-modal consistency is a key factor in multimodal fact-checking—image-based questions must align with textual claims without introducing unintended biases or assumptions.

\noindent\textbf{LLM-Based vs. Human Evaluation}. To ensure consistency, we employ GPT-4o as an automated evaluator, scoring FCQs based on predefined criteria such as logical structure, factual precision, and investigative depth. However, LLM-based evaluations may still miss nuanced contextual ambiguities that a human fact-checker would recognize, such as misleading phrasing or assumptions embedded in a question.

To validate the reliability of the LLM-based assessment, we conducted a human agreement study, comparing GPT-4o’s evaluation results with expert annotations across datasets in Table~\ref{tab:correlation_scores_evaluation}. The goal was to determine the degree of alignment between human and LLM judgments, rather than integrating both assessments into a single process. 

Our findings indicate that while LLMs are effective at systematically evaluating FCQs, human reviewers provide valuable qualitative insights, particularly in identifying question formulation errors that could lead to misinformation rather than prevent it. 

This subtle difference can impact both retrieval accuracy and the framing of fact-checking results.

\noindent\textbf{Insights from the Evaluation Process}. 
\begin{itemize}
    \item \textbf{Text-based FCQs generally receive higher relevance scores than image-based FCQs.} This discrepancy suggests that LLMs have a better grasp of linguistic verification than visual reasoning, which remains an open challenge in multimodal misinformation detection.
    \item \textbf{Human annotators tend to be stricter in rejecting vague or broad FCQs.} LLM-based evaluations show slightly higher acceptance rates for questions that are loosely related to the claim but lack clear fact-checking intent.
    \item \textbf{Context-aware evaluation is critical.} Without access to real-world updates, an FCQ might appear factually valid but be outdated or misleading in light of new developments. This highlights the importance of external knowledge retrieval in automated fact-checking pipelines.
\end{itemize}

By comparing human and LLM-based assessments, our study confirms that GPT-4o produces highly relevant FCQs with near-human accuracy. However, human reviewers remain essential in refining question design and identifying subtle logical inconsistencies that automated evaluations may overlook.
\begin{table*}[t]
\centering
\resizebox{\textwidth}{!}{%
\begin{tabular}{l l}
\toprule[1.5pt]
\textbf{Criteria} & \textbf{Definition} \\ 
\midrule

Critical Thinking and Skepticism & Challenges assumptions, probes deeper into claims, avoids taking information at face value.\\ 
Analytical Depth & Breaks down complex statements into verifiable components.\\ 
Systematic Approach & Follows a structured methodology in assessing sources, claims, and evidence.\\ 
Precision \& Specificity & Clear, direct, and free from vague or overly broad wording. \\ 
Factual Accuracy & Focuses on verifying evidence, checking primary sources, and detecting misinformation.\\ 
Logical Consistency & Identifies contradictions, misleading narratives, or inconsistencies. \\ 
Source Credibility \& Bias Detection & Evaluates the reliability of cited sources and potential biases. \\ 
Context Awareness & Considers the broader context surrounding the claim.\\ 
Comparative Thinking & Encourages cross-referencing with established facts or alternative perspectives. \\ 
Repeatability and Objectivity & Can be applied consistently across different cases without personal bias.\\ 

\toprule[1.5pt]
\end{tabular}%
}
\caption{Criteria for assessing the accuracy, credibility, and reliability of FCQs.}
\label{tab:tabel_criteria}
\end{table*} 

\subsection{Dataset Descriptions and Details}
\label{dataset}

To assess the effectiveness of LLM-generated fact-checking questions in multimodal misinformation detection, we utilize three benchmark datasets: MMFakeBench, DGM4, and Factify. These datasets encompass a wide range of real and manipulated image-text pairs, enabling a comprehensive evaluation of textual, visual, and cross-modal inconsistencies. Each dataset provides a distinct annotation scheme, capturing various types of misinformation, from textual distortions to manipulated images and multimodal inconsistencies.

\noindent\textbf{MMFakeBench Dataset}. This dataset serves as a benchmark for multimodal misinformation detection. It categorizes misinformation into three primary types:

\begin{itemize}
    \item \textbf{Textual Veracity Distortion (TVD)}: Fake or misleading textual claims.
    \item \textbf{Visual Veracity Distortion (VVD)}: Manipulated or AI-generated images.
    \item \textbf{Cross-Modal Consistency Distortion (CMM)}: Mismatches between text and images.
\end{itemize}

Each sample in MMFakeBench is annotated based on:
\begin{itemize}
    \item Whether the claim text is factually correct.
    \item Whether the accompanying image has been manipulated.
    \item Whether the text-image pair is consistent or inconsistent.
\end{itemize}

The dataset provides a structured framework to evaluate misinformation detection across multiple manipulation types, incorporating diverse real-world scenarios.

\noindent\textbf{DGM4 Dataset}. The DGM4 (Grounding Multi-Modal Media Manipulation) dataset is a large-scale collection of manipulated and real news samples focusing on human-centric content. It contains approximately 230,000 samples, distributed as follows:

\begin{itemize}
    \item \textbf{77,426} pristine (real) image-text pairs.
    \item \textbf{152,574} manipulated samples, generated using:
    \begin{itemize}
        \item \textbf{Face Swap (FS)}: 66,722 samples.
        \item \textbf{Face Attribute Manipulation (FA)}: 56,411 samples.
        \item \textbf{Text Swap (TS)}: 43,546 samples.
        \item \textbf{Text Attribute Manipulation (TA)}: 18,588 samples.
        \item \textbf{Mixed Manipulation Pairs}: 32,693 samples combining text and image edits.
    \end{itemize}
\end{itemize}

\noindent\textbf{Factify Dataset}. This dataset is a multimodal fact verification dataset containing 50,000 samples collected from Twitter and online news sources in the United States and India. Each sample consists of:
\begin{itemize}
    \item \textbf{Claim text}: A short statement, often extracted from tweets.
    \item \textbf{Claim image}: The corresponding image that either supports or contradicts the claim.
    \item \textbf{OCR text}: Extracted text from the claim image.
    \item \textbf{Document text}: A news article serving as supporting evidence.
    \item \textbf{Document image}: An image from the referenced news article.
\end{itemize}

Samples in Factify are categorized into five classes:
\begin{itemize}
    \item \textbf{Support\_Text}: The claim text is supported by the document text, but images are dissimilar.
    \item \textbf{Support\_Multimodal}: Both the claim text and image match the document text and image.
    \item \textbf{Insufficient\_Text}: The document does not provide enough textual evidence to support or refute the claim.
    \item \textbf{Insufficient\_Multimodal}: The document image matches the claim image, but the text lacks confirmation.
    \item \textbf{Refute}: The document contradicts both the claim text and the claim image.
\end{itemize}

The dataset provides a benchmark for multimodal fact verification, leveraging both textual and visual evidence to assess claim veracity.

\begin{figure*}[t]
	\centering
	\includegraphics[width=1\textwidth]{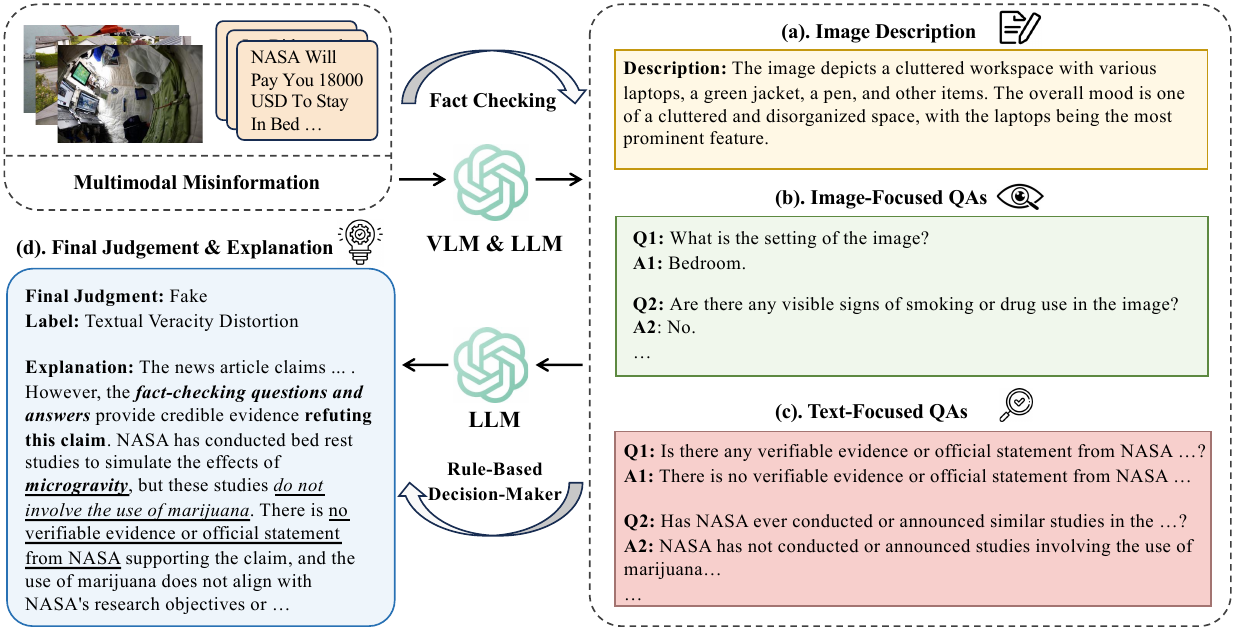}
	\caption{The overview pipeline of our \textsc{Lrq-Fact} framework consists of four key components: (a) Image Description, which provides detailed contextual descriptions of the image; (b) Visual FCQs, aimed at assessing the accuracy of the visual content; and (c) Textual FCQs, which detect textual inaccuracies, contradictions, or unsupported claims. Finally, all the gathered information is synthesized in (d) the Final Judgment \& Explanation module, where a rule-based decision-maker generates both the prediction results and comprehensive explanations.}
	\label{fig:model_pipeline}
\end{figure*}

\begin{figure*}[!t]
    \centering

    \begin{tcolorbox}[colback=yellow!10!white,colframe=gray!75!black,title={\textsc{Image Description Prompt}:}]
    \textit{\small Please provide a detailed and comprehensive description of the image shown. Focus on identifying all visible elements including objects, people, setting, and any interactions or actions taking place. Describe the colors, textures, mood, and any other notable aspects that contribute to the overall context and significance of the image.}
    \end{tcolorbox}
    \vspace{-0.3cm}
    \caption{Structured prompt to generate detailed image descriptions.}
    \label{fig:prompt_img_description}
    
\end{figure*}

\newpage


\begin{figure*}[t]
    \centering
    
    \begin{tcolorbox}[colback=yellow!10!white,colframe=gray!75!black,title=\text{\textsc{Visual Questions Prompt}:}]
    {\small Given the following news article [\textbf{news text}], generate up to [\textbf{number of questions}] questions that are directly based on the news article and are designed to explore visual elements that could be present in an image related to the article.
}\\
    
    \textbf{Instructions for Question Generation:}\\
    \textit{\small Focus on generating questions that are directly relevant to the news article and the visual elements that could be present in an image. The questions should examine visible interactions, settings, actions, text, symbols, and specific objects mentioned in the article. Additionally, include questions that assess the authenticity of the image, such as whether it could have been AI-generated or contains any unusual or suspicious elements. 
}\\

    \textbf{Avoid the following in your Questions:}\\
    \textit{\small - Do not mention any names.\\
    - Do not ask questions about identification .\\
    - Do not ask about personal details.\\
    - Do not ask compound questions in a single sentence.}\\

    \textbf{Example Questions:}\\
    \textit{\small 1. What event is depicted in this image?\\
    2. How are the people in the image interacting?\\
    3. Is the person in the image performing [\textbf{action from article}]?\\
    4. What are the technical aspects or tools used to create this image?\\
    5. What emotions does this image evoke?\\
    6. What are the main objects or elements visible in this image?\\
    7. What unusual elements in the image might suggest digital manipulation or artificial creation?\\
    ...}
    \tcblower
    \textbf{Questions:}
    \textit{\small 1. , 2. , ...}

    \end{tcolorbox}
    \vspace{-0.3cm}
    \caption{Structured prompt to generate relevant visual questions.}
    \label{fig:prompt_img_q}
    \vspace{-0.3cm}
\end{figure*}


\begin{figure*}[!t]
    \centering
    \begin{tcolorbox}[colback=yellow!10!white,colframe=gray!75!black,title=\text{\textsc{Visual Answers Prompt}:}]
    {\small You are an advanced AI model with access to a vast repository of knowledge and the capability of answering image questions. Your task is to answer the following questions [\textbf{generated questions}] based on the image [\textbf{image}]. While a news article [\textbf{news text}] is provided for context, you must answer the questions solely based on the image and not refer to the article's content.}\\
    
    \textbf{Instructions for Answer Generation:}\\
    \textit{\small - Provide accurate, clear, and concise answers to each question.\\
    - Your responses should be based entirely on the image.\\
    - Do not reference or rely on the content of the provided news article when forming your answers.\\
    - Each answer should be directly relevant to the question asked.
    }\\

    \textbf{Avoid the following in your Answers:}\\
    \textit{\small - Provide accurate, clear, and concise answers to each question.\\
    - Your responses should be based entirely on the image.\\
    - Do not reference or rely on the content of the provided news article when forming your answers.\\
    - Each answer should be directly relevant to the question asked.
    }
    
    \tcblower
    \textbf{Answers:}
    \textit{\small 1. , 2. , ...}

    \end{tcolorbox}
    \vspace{-0.3cm}
    \caption{Structured prompt to generate answers for the visual questions.}
    \label{fig:prompt_img_a}
    \vspace{-0.3cm}
\end{figure*}


\begin{figure*}[!t]
    \centering
    
    \begin{tcolorbox}[colback=yellow!10!white,colframe=gray!75!black,title=\text{\textsc{Textual Questions Prompt}:}]    
    {\small Given the following news article [\textbf{news text}], analyze the text and formulate up to [\textbf{number of questions}] questions that probe the accuracy and verifiability of the information contained in the article. These questions should be designed to identify potential inaccuracies or areas that can be confirmed or challenged based on general knowledge or the text itself.}\\
    
    \textbf{Instructions for Question Generation:}\\
    \textit{\small Focus on generating high-quality, fact-checking questions that can be answered directly through general knowledge that an LLM might possess. Identify and question significant factual claims, examine dates, locations, names, and other data mentioned in the article, and challenge any assumptions. The goal is to produce questions that facilitate direct verification of the facts stated in the article.}\\

    \textbf{Aim to Generate:}\\
    \textit{\small - Questions that challenge the accuracy of specific claims made in the article and can be answered based on general knowledge.\\
    - Questions that explore potential inconsistencies or contradictions within the article's content.\\
    - Questions that assess the logical coherence and factual basis of the article's claims.}\\\\
    \textbf{Avoid asking for:}\\
    \textit{\small- Information requiring external sources or verification beyond general knowledge.\\
    - Speculative or opinion-based questions.}\\

    \textbf{Example Questions:}\\
    \textit{\small 1. Does the description of the ``meeting between world leaders on March 5th'' align with the known schedule of diplomatic events for that time?\\
    2. Is the account of ``a large protest taking place in front of City Hall'' consistent with known reports of protests in that area during the stated period?\\
    3. Does the timeline of ``economic sanctions being imposed after the incident'' logically follow the typical process for such actions?\\
    4. Are the historical events referenced, such as ``the financial crisis of 2008'', accurately portrayed in the article?\\
    ...}
    \tcblower
    \textbf{Questions:}
    \textit{\small 1. , 2. , ...}

    \end{tcolorbox}
    \vspace{-0.3cm}
    \caption{Structured prompt to generate relevant textual questions.}
    \label{fig:prompt_text_q}
    \vspace{-0.3cm}
\end{figure*}


\begin{figure*}[!t]
    \centering
    
    \begin{tcolorbox}[colback=yellow!10!white,colframe=gray!75!black,title=\text{\textsc{Textual Answers (w/o Evidence) Prompt}:}]    
    {\small You are an advanced AI model with access to a vast repository of knowledge. Your task is to answer the following questions [\textbf{generated questions}] based on your built-in knowledge. While a news article [\textbf{news text}] is provided for context, you must answer the questions solely based on your own knowledge and not refer to the article's content.}\\
    
    \textbf{Instructions for Answering:}\\
    \textit{\small Provide accurate, clear, and concise answers to each question. Your responses should be based entirely on your general knowledge and the information you have learned. Do not reference or rely on the content of the provided news article when forming your answers. Each answer should be factually correct and directly relevant to the question asked.}
   
    \tcblower
    \textbf{Answers:}
    \textit{\small 1. , 2. , ...}

    \end{tcolorbox}
    \vspace{-0.3cm}
    \caption{Structured prompt to generate answers based on llm-knowledge for the relevant textual questions.}
    \label{fig:prompt_text_a}
    \vspace{-0.3cm}
\end{figure*}


\begin{figure*}[!t]
    \centering
    
    \begin{tcolorbox}[colback=yellow!10!white,colframe=gray!75!black,title=\text{\textsc{Textual Answers (w/Evidence) Prompt}:}]    
    {\small You are an advanced AI tasked with evaluating the authenticity of a news article. Your task is to answer the following questions [\textbf{generated questions}] based on the provided factual document [\textbf{evidence}]. While a news article [\textbf{news text}] is provided for context, you must answer the questions solely based on the provided factual document and not refer to the article's content.}\\
    
    \textbf{Instructions for Answering:}\\
    \textit{\small Provide accurate, clear, and concise answers to each question. Your responses should be based entirely on the provided factual document. If there was no factual answer for the question use your built-in knowledge to answer the question. Do not reference or rely on the content of the provided news article when forming your answers. Each answer should be directly relevant to the question asked.
}
   
    \tcblower
    \textbf{Answers:}
    \textit{\small 1. , 2. , ...}

    \end{tcolorbox}
    \vspace{-0.3cm}
    \caption{Structured prompt to generate answers based on factual evidence for the relevant textual questions.}
    \label{fig:prompt_text_a_RAG}
    \vspace{-0.3cm}
\end{figure*}


\begin{figure*}[!t]
    \centering
    
    \begin{tcolorbox}[colback=yellow!10!white,colframe=gray!75!black,title=\text{\textsc{Questions Quality Prompt}:}]    
    {\small You are an advanced AI tasked with evaluating the authenticity of a news article and its accompanying image. Your objective is to determine whether the provided questions [\textbf{generated questions}] effectively assess the accuracy, credibility, and reliability of the news article text.}\\
    
    \textbf{Instructions:}\\
    \textit{\small Assess each question based on the following expert fact-checking criteria:\\
    1. Critical Thinking and Skepticism: Does the question challenge assumptions, probe deeper into claims, and avoid taking information at face value?\\
    2. Analytical Depth: Does it break down complex statements into verifiable components?\\
    3. Systematic Approach: Does it follow a structured methodology in assessing sources, claims, and evidence?\\
    4. Precision \& Specificity: Is it clear, direct, and free from vague or overly broad wording?\\
    5. Factual Accuracy: Does it focus on verifying evidence, checking primary sources, and detecting misinformation?\\
    6. Logical Consistency: Does it help identify contradictions, misleading narratives, or inconsistencies?\\
    7. Source Credibility \& Bias Detection: Does it evaluate the reliability of cited sources and potential biases?\\
    8. Context Awareness: Does it consider the broader context surrounding the claim?\\
    9. Comparative Thinking: Does it encourage cross-referencing with established facts or alternative perspectives?\\
    10. Repeatability \& Objectivity: Can the question be applied consistently across different cases without personal bias?}\\

    \textbf{Rating Scale:}\\
    \textit{\small - Relevant: The question is precise, well-structured, and effectively assesses factual accuracy, credibility, and logical consistency.\\
    - Irrelevant: The question is vague, lacks depth, or fails to critically probe the credibility and factuality.}\\
    
    \tcblower
    \textbf{Answers:}\\
    \textit{\small Q1: [Relevant or Irrelevant]\\
        Q2: [Relevant or Irrelevant]\\
        ...}

    \end{tcolorbox}
    \vspace{-0.3cm}
    \caption{Structured prompt to evaluate the quality of generated questions.}
    \label{fig:prompt_question_quality}
    \vspace{-0.3cm}
\end{figure*}


\begin{figure*}[!t]
    \centering
    
    \begin{tcolorbox}[colback=yellow!10!white,colframe=gray!75!black,title=\text{\textsc{Rule-Based Decision-Maker Prompt}:}]
    {Your objective is to determine whether the article and image are real or fake by analyzing the following information:}\\
    \textit{\small
    1. News Article Text: [\textbf{news text}]\\\\
    2. Image Description: [\textbf{image description}]\\
    Note: This description helps verify consistency with the news text. It is generally reliable but may contain minor discrepancies, such as using different terms like ``ocean'' instead of ``water''.\\\\
    3. Generated Visual Questions and Answers: [\textbf{generated visual FCQs}]\\
    Note: These answers were generated by an AI and may contain mistakes, such as incorrect details regarding locations, names, dates, or objects. They might also incorrectly suggest that the image has been manipulated or is AI-generated. If the answers suggest manipulation or that the image is AI-generated, this should have very low effect on your final decision, especially if the image description and news article text do not contain such indications.\\\\
    4. Generated Textual Questions and Answers: [\textbf{generated textual FCQs}]\\
    Note: These are based on the knowledge of GPT4-O, which is generally reliable but prone to hallucinations or contradictions with other provided information.}\\
    
    \textbf{Instructions:}\\
    \textit{\small To make an accurate judgment of the multimodal misinformation, please follow these steps:\\
    \textbf{Step 1.} Is there any credible objective evidence refuting the news description? If yes, assign the label: Textual Veracity Distortion. If no, continue to Step 2.\\
    \textbf{Step 2.} Is there any credible objective evidence refuting the news image? If yes, assign the label: Visual Veracity Distortion. If no, continue to Step 3.\\
    \textbf{Step 3.} Does the news caption match the content of the news image? If no, assign the label: Mismatch. If yes, and none of the above applies, assign the label: Real.}\\

    \textbf{Additional Guidelines:}\\
    \textit{\small
    1. Assess Overall Consistency: ...\\
    2. Examine Details: ...\\
    3. Analyze Facial Expressions and Body Language: ...\\
    4. Identify Unrealistic Elements: ...\\
    5. Cross-Modal Consistency: ...\\
    6. Final Judgment: ...\\
    7. Select the Most Relevant Label: ...\\
    8. Provide a Detailed Explanation: ...\\}
    
    \textbf{Example Output:}\\
    \small{\textbf{1. Final Judgment:} \textit{Fake}\\
    \textbf{2. Label:} \textit{Visual Veracity Distortion}\\
    \textbf{3. Explanation:} \textit{The image description mentions a ``cat with pink eyes'', which is highly unnatural and suggests the image is AI-generated. Additionally, ....}}
    \tcblower
    \small{
    \textbf{1. Final Judgment:} [Real or Fake]\\
    \textbf{2. Label:} [Select one: Textual Veracity Distortion, Visual Veracity Distortion, Mismatch, Real]\\
    \textbf{3. Explanation:} [Provide your explanation here]}

    \end{tcolorbox}
    \vspace{-0.3cm}
    \caption{Structured prompt to make the final decisions and provide an explanation.}
    \label{fig:prompt_judge}
    \vspace{-0.3cm}
\end{figure*}

\section{Acknowledgment of AI Assistance in Writing and Revision}
We utilized ChatGPT-4 for revising and enhancing sections of this paper.

\end{document}